\documentclass[11pt]{article}
\usepackage[utf8]{inputenc}
\usepackage[T1]{fontenc}
\usepackage{nodalida2025}
\usepackage{times}
\usepackage{url}
\usepackage{latexsym}
\usepackage{booktabs}
\usepackage{graphicx}

\usepackage{hyperref}

\aclfinalcopy 

\title{Hotter and Colder: A New Approach to Annotating Sentiment, Emotions, and Bias in Icelandic Blog Comments}

\author{
    Steinunn Rut Friðriksdóttir \\
    University of Iceland\\
    {\tt srf2@hi.is} \\\And
    Dan Saattrup Nielsen \\
    The Alexandra Institute\\
    {\tt dan.nielsen@alexandra.dk} \\\And
    Hafsteinn Einarsson \\
    University of Iceland \\
    {\tt hafsteinne@hi.is}
}

\date{}

\begin{document}
\maketitle
\begin{abstract}
This paper presents Hotter and Colder, a dataset designed to analyze various types of online behavior in Icelandic blog comments. Building on previous work, we used GPT-4o mini to annotate approximately 800,000 comments for 25 tasks, including sentiment analysis, emotion detection, hate speech, and group generalizations. Each comment was automatically labeled on a 5-point Likert scale. In a second annotation stage, comments with high or low probabilities of containing each examined behavior were subjected to manual revision. By leveraging crowdworkers to refine these automatically labeled comments, we ensure the quality and accuracy of our dataset resulting in 12,232 uniquely annotated comments and 19,301 annotations. Hotter and Colder provides an essential resource for advancing research in content moderation and automatically detectiong harmful online behaviors in Icelandic. We release both the dataset\footnote{\url{https://repository.clarin.is/repository/xmlui/handle/20.500.12537/352}} and annotation interface\footnote{\url{https://github.com/icelandic-lt/annotation_if_sentiment}}.
\end{abstract}

\section{Introduction}
The rapid growth of online communication platforms has led to an increase in harmful behaviors and, subsequently, an increased need for content moderation~\cite{mathew2019spread}. Inappropriate comments targeted at specific individuals or groups of people can even go so far as qualifying as hate speech, but more subtle ways of spreading these prejudiced ideas may, for instance, include fear speech, where attempts are made to incite fear about a target community~\cite{saha2023rise}. Recent work has focused on detecting these toxic behaviors automatically, thereby lessening the cost and workload for human moderators (see~\citet{dehghan2024evaluating},~\citet{nagar2023towards} and~\citet{mittal2023detecting} for instance).  

This paper addresses limitations in previous work on sentiment analysis in Icelandic~\cite{fridriksdottir-etal-2024-ice}, using a new methodology to improve class imbalance and low annotator agreement in some tasks. Our approach first uses GPT-4o mini to analyze approximately 800,000 Icelandic blog comments across 25 tasks, including sentiment analysis, emotion detection, hate speech detection, and group generalizations. For most tasks, we employ focused binary annotation, targeting only the extreme cases (highly likely or highly unlikely to exhibit the behavior), rather than using rating scales which have been shown to present challenges in maintaining consistent annotation quality~\cite{kiritchenko-mohammad-2017-best}. The exception is sentiment analysis, where we maintain the standard negative, neutral, and positive categories.


This targeted approach allows us to efficiently identify rare but important cases (the proverbial needles-in-a-haystack) such as hate speech comments, which would be resource-intensive to locate through random sampling as used in previous work. To ensure dataset quality, we then employ crowd workers to manually verify the model's predictions, focusing particularly on comments flagged as highly likely or highly unlikely to contain problematic content. This human verification step is crucial for maintaining accuracy and creating a high-consensus dataset.


Our contributions are as follows:
\begin{itemize}
    \item We present Hotter and Colder, a dataset of 12,232 Icelandic blog comments annotated for 25 tasks including sentiment, emotions, hate speech, and group generalizations
    \item We introduce a two-phase annotation methodology combining GPT-4o mini silver labels with targeted human verification to address class imbalance and improve annotation agreement
    \item We release both the annotated dataset and annotation platform to support research in content moderation for low-resource languages\footnote{[links redacted]}
\end{itemize}

\section{Methodology}
Our approach combines AI and human efforts in a two-phase annotation process designed to create a high-quality dataset for tasks where the phenomena of interest are often rare. This scarcity poses a significant challenge for dataset creation - random sampling would require extensive human annotation effort to find sufficient positive examples while focusing only on suspected positive cases could bias the dataset. Our methodology aims to balance these concerns by using AI to efficiently identify potential cases across the full spectrum, followed by targeted human verification.


In the first phase (silver labeling), an LLM analyzes a large dataset of comments. For this initial screening, we use GPT-4o mini with a prompt designed for structured output (see Section~\ref{sec:silver}). While the model was instructed to consider itself an expert in Icelandic blog analysis to maintain consistent task framing across annotations, we acknowledge this is a common but debatable prompting practice that warrants further investigation. For all tasks except sentiment analysis, the LLM uses a 5-point scale for labeling to capture nuanced assessments.

In the second phase (gold labeling), human annotators review selected comments, focusing primarily on those the LLM rated at the extremes of the scale (1 or 5). This design choice reflects our priority of establishing a foundational dataset with clear, agreed-upon examples of each phenomenon. While this approach may not capture all nuanced edge cases, it serves several important purposes: (1) it enables efficient identification of clear positive examples for rare phenomena, (2) it helps establish reliable baseline annotations for model evaluation, and (3) it aligns with findings that human annotators achieve higher agreement on clear cases~\cite{kiritchenko-mohammad-2017-best}. We acknowledge this as a limitation - future work should explicitly target borderline cases to improve model robustness.

Human annotators perform binary (yes/no) annotations\footnote{Hick's law states that increasing the number of choices will increase the time it takes a person to make a decision logarithmically~\cite{hick1952rate}.} for a single task at a time to reduce task switching fatigue. The simplified binary choice for humans, compared to the LLM's 5-point scale, reflects our focus on identifying clear instances while acknowledging that intermediate cases may require more nuanced future investigation.

This method of using a language model to identify potential candidates for gold labeling builds on established practices. For instance, when compiling their GoEmotions dataset, \citet{demszky-etal-2020-goemotions} used a BERT-based model to filter out comments that contained high levels of neutrality, leaving the more emotional comments for humans to annotate.



\subsection{Silver Labeling Phase}\label{sec:silver}
To automate the initial labeling process, we created a prompt for the AI model that instructed the model to perform all of the 25 annotation tasks on a given blog comment in Icelandic\footnote{\url{https://gist.github.com/Haffi112/8813b738637fc9a678f524fdf9b5a5d9}}. The prompt included a JSON schema that instructed the model on how to label a given comment. The context provided to the model also included the previous comments and the beginning of the blog post on which the comments were posted. We used strictly structured outputs to guarantee that the GPT-4o mini model always labeled each comment for each of the 25 tasks and to make sure that it could only output values that aligned with the Likert scale\footnote{See information on OpenAI's website \href{https://openai.com/index/introducing-structured-outputs-in-the-api/}{here}.}. 

\subsection{Data Selection}
Following the previous work of~\citet{fridriksdottir-etal-2024-ice}, the blog comments used in this work all derive from the Icelandic blog platform \texttt{blog.is}. As one of the oldest and still active blogging platforms in Iceland, this website offers a valuable collection of online communication, generating a wide range of debates between people with different perspectives, which is particularly useful for our purposes. However, it should be noted that the gender distribution of the site's users appears to be quite skewed. \texttt{Blog.is} has no obvious demographics accessible for users. In his master thesis, however, \citet{asmundssonanalyzing} used a heuristic approach to determine the gender of the users based on their patronyms (traditionally, women's last names in Icelandic end with \textit{dóttir} (e. \textit{daugther}) and men's last names end with \textit{son}). Similarly, we observed that out of 24,193 unique author names, 2,374 ended in ``dóttir'', 7,539 ended in ``son'' and 14,280 user names did not match these endings.

\subsection{Task Overview}
The LLM was provided with the context of the blog post, previous comments, and the specific comment to be analyzed. The system prompt for the model was ``You are an expert at analyzing Icelandic blog comments. Analyze the last comment shown and provide insights based on the given schema.'' For a given input, the model generated its analysis according to a predefined JSON schema, ensuring consistency across all evaluated comments.

The analysis began with an overall sentiment classification (positive, negative, or neutral) of each comment. The LLM then evaluated a wide range of attributes, including toxicity, politeness, hate speech, social acceptability in various contexts, emotional content, sarcasm, constructiveness, encouragement, sympathy, trolling behavior, mansplaining, and group generalizations. For hate speech, the model identified specific target groups and aggression levels when present. The analysis of group generalizations included assessing sentiment, factual validity, and whether the mentioned groups were marginalized.

Most attributes were rated on a 5-point Likert scale, where 1 indicated strong disagreement and 5 indicated strong agreement with the presence or intensity of the attribute\footnote{Rubric: 1 - Strongly Disagree, 2 - Disagree, 3 - Neither Agree nor Disagree, 4 - Agree, 5 - Strongly Agree}. For some attributes, such as sentiment (``positive'', ``neutral'', ``negative'') and gender (``male'', ``female'', ``non-binary'', ``n/a''), predefined categories were used instead.


We selected our emotion categories based on the foundational work of \citet{ekman1992argument,ekman1988universality}, who identified seven basic emotions that appear to be universal across cultures: fear, happiness, sadness, surprise, disgust, anger, and contempt. To this set, we added indignation as it represents a distinct social emotion particularly relevant to online discourse and content moderation. Social acceptability was assessed across various contexts, including conversations with strangers, acquaintances, and close friends, in educational settings with different age groups, and in parliamentary speeches.

The LLM also inferred the author's gender and we further performed a majority vote over all annotations of a given username to assign a gender to the author's name. We note that gender inference in online spaces presents significant challenges. While traditional Icelandic naming conventions can provide gender cues through patronymic suffixes (-son/-dóttir), we acknowledge several important limitations in our approach to gender inference:
\begin{enumerate}
    \item Users may choose pseudonyms that do not reflect their actual gender, particularly given documented patterns of gender-based harassment online.
    \item The relationship between usernames and actual gender identity is complex and cannot be reliably determined through automated analysis.
    \item Some users may intentionally obscure their gender or choose gender-neutral identifiers.
\end{enumerate}
We emphasize that the inferred gender labels should be treated as approximations of perceived rather than actual gender, particularly in analyses of gendered interaction patterns like mansplaining. Future work should explore alternative approaches to studying gendered communication patterns that do not rely on automated gender inference.

\subsection{Human Annotation Process}
To evaluate Icelandic blog comments, we developed a comprehensive annotation scheme covering various aspects of online discourse. Human annotators were provided with detailed instructions in Icelandic, emphasizing that their personal judgment was crucial and that there were no strictly right or wrong answers. Annotators were instructed to base their decisions on the content of the comments rather than the authors' names, of which only initials and inferred gender were provided.

For most tasks, annotators were asked to make binary decisions (yes/no) about whether a comment exhibited specific characteristics. The exception was sentiment analysis, which used a three-way classification. Annotators could view preceding comments and the original blog post for context, although some images were no longer available. They were also given the option to skip annotation for comments containing minimal information or those in languages other than Icelandic.

\subsubsection{Sentiment Analysis}
Following the approach of \citet{wankhade2022survey}, we conducted sentiment analysis at the comment level. Annotators classified each comment as positive, negative, or neutral based on their personal interpretation. Positive sentiment was defined as expressing approval, happiness, satisfaction, or optimism. Negative sentiment indicated dissatisfaction, criticism, anger, or disappointment. Neutral sentiment was characterized by a lack of strong emotion or a balanced view, often seen in informational or factual statements.

\subsubsection{Toxicity}
We adopted the definition of toxicity in online discussions from \citet{klein2024medium}, describing it as behavior that is rude, disrespectful, or unreasonable, potentially making users feel unwelcome or discouraged from participating in the discussion. Annotators were instructed to identify comments containing insults, aggressive language, or content likely to incite conflict. This approach acknowledges the potential of toxic comments to disrupt constructive dialogue and decrease user engagement, as observed in studies of online forums \citep{young2024responding}.

\subsubsection{Hate Speech}
Our hate speech annotation scheme was based on \citet{basile-etal-2019-semeval} and aligned with Article 233 (a) of the Icelandic penal code, an approach also used by \citet{fridriksdottir-etal-2024-ice}. Annotators identified comments containing threats, defamation, or denigration based on protected characteristics such as nationality, color, race, religion, sexual orientation, disabilities, or gender identity.

\subsubsection{Social Acceptance}
To gauge social acceptability, annotators evaluated whether it would be appropriate to make the comment in question in various real-life contexts. These included interactions with strangers, acquaintances, and close friends, as well as in educational settings (for both young children and teenagers) and in parliamentary speeches. This multi-context approach allowed for a nuanced understanding of perceived social norms across different situations.

\subsubsection{Emotion Detection}
Our emotion detection task was inspired by the work of \citet{fridriksdottir-etal-2024-ice} and \citet{demszky-etal-2020-goemotions}. We simplified the task by asking annotators to detect the presence of a single emotion at a time in a binary fashion. In other words, to answer whether or not a comment contained the given emotion. The emotions included were based on basic emotions identified by \citet{ekman1992argument} and \citet{ekman1988universality}: fear, happiness, sadness, surprise, disgust, anger, and contempt. We also included indignation.

\subsubsection{Sarcasm}
Following the approach of~\citet{ptacek-etal-2014-sarcasm}, we asked the annotators to label whether a given comment was sarcastic or ironic. In Icelandic, there is a tendency to lump these two meanings together in one (ice.\ \textit{kaldhæðni}).

\subsubsection{Constructiveness}
We employed a simplified version of the annotation scheme from \citet{kolhatkar2020classifying}, asking annotators to determine whether comments were constructive. This binary classification focused on identifying comments that provided useful feedback or contributed positively to the discussion.

\subsubsection{Encouragement and Sympathy}
Inspired by \citet{sosea-caragea-2022-ensynet}, we asked annotators to identify encouragement and sympathy in comments in a binary fashion. Encouragement was defined as inspirational words or support and sympathy was defined to be compassion, pity, or understanding of the situation of another person.

\subsubsection{Additional Annotations}
We included several other classification tasks to capture various aspects of online discourse:

\textbf{Politeness:} Annotators assessed whether comments were polite, providing a measure of civility in online interactions.

\textbf{Trolling:} Following the definition used by \citet{fridriksdottir-etal-2024-ice}, we asked annotators to identify comments that were intentionally provocative, offensive, or off-topic, aimed at eliciting strong emotional responses or disrupting normal discussion.

\textbf{Mansplaining:} The term has been defined by \citet{bridges2017gendering} as ``a man explaining something to a woman in a tone perceived as condescending,'' but has since been expanded to cover a broader range of communicative behaviors~\citep{smith2022well}. Annotators were instructed to identify instances where comments exhibited unsolicited, patronizing explanations based on the assumption that the recipient is ignorant. Key characteristics of mansplaining include:
\begin{itemize}
\item Persistence even when the recipient demonstrates expertise.
\item Maintenance of an oversimplified approach.
\item Unwarranted confidence, sometimes even when factually incorrect.
\end{itemize}
While mansplaining can occur between individuals of any gender, annotators were instructed to use the label only for male-to-female interactions. The gendered term highlights the frequency of this dynamic in male-female conversations, particularly in fields where women may have equal or superior expertise. This annotation task aimed to reveal ongoing societal assumptions about gender, knowledge, and competence, illustrating how gender-based power dynamics continue to shape interpersonal and professional communications.

\textbf{Group Generalizations:} Annotators were asked to identify comments containing broad, often oversimplified statements about entire groups of people. These generalizations could be based on characteristics such as race, gender, nationality, or political views. Importantly, annotators were instructed to note that these generalizations could be positive, negative, or neutral in nature. This task aimed to capture instances where comments reflected biases, stereotypes, or assumptions about groups, providing insight into how these generalizations manifest in online discourse.

\subsection{Agreement Measures}
To evaluate annotation quality and reliability, we employed multiple agreement metrics. For tasks with two or more annotations per comment, we calculated pairwise agreement (PA) as the proportion of agreeing annotation pairs across all possible pairs. For assessing inter-annotator reliability, we utilized Krippendorff's alpha (K's $\alpha$), which accounts for chance agreement and can handle missing data — a common occurrence in crowdsourced annotations. To evaluate the GPT-4o mini's performance against human judgments, we computed Cohen's kappa (C's $\kappa$) between the model's predictions and the human consensus labels that were computed through a majority vote (examples with ties were dropped). For the sentiment analysis task, which involved three-way classification, we adapted these measures to account for the additional category whilst maintaining the same computational framework.

\subsection{Annotation Interface}
The annotation interface was designed to facilitate efficient and accurate labeling of blog comments while providing contextual information to annotators. The interface presents one comment at a time, along with metadata such as the author's initials, inferred gender, and timestamp. To enhance context, annotators can optionally view the full blog post and previous comments in the thread where the same type of metadata is shown for each author. Tasks are presented sequentially, with clear instructions and the option to skip comments when necessary. To maintain engagement and provide feedback, the interface incorporates gamification elements such as progress tracking and achievement badges.

To ensure data quality, the interface implements several key features. First, it allows annotators to review task-specific guidelines at any point during the annotation process. Second, the interface offers an optional real-time feedback mechanism that compares human annotations to predictions from GPT-4o-mini, though annotators are explicitly instructed to rely on their own judgment rather than attempting to match the model's output. This design balances the need for comprehensive contextual information with the goal of maintaining annotator focus and efficiency throughout the task.

\begin{figure*}[t] 
    \centering
    \includegraphics[width=\textwidth]{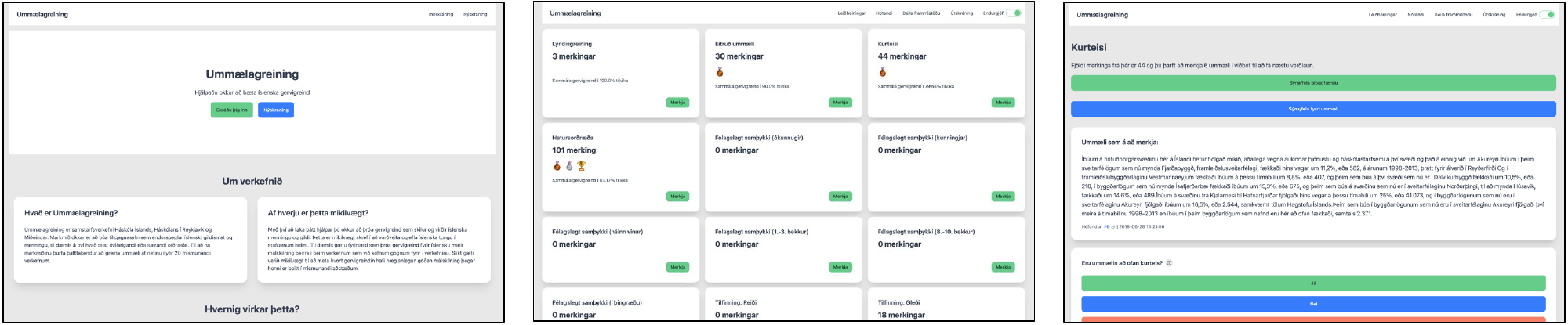}
    \caption{Key components of the annotation platform: (left) The landing page introducing the project and its importance; (middle) The task overview dashboard displaying user progress and available annotation tasks; (right) An example of a specific annotation task (politeness assessment) showing the comment to be annotated, contextual information, and annotation options.}
    \label{fig:interface}
\end{figure*}

\section{Results}

\subsection{Distribution of AI labels}
\begin{figure}
    \centering
    \includegraphics[width=1.0\linewidth]{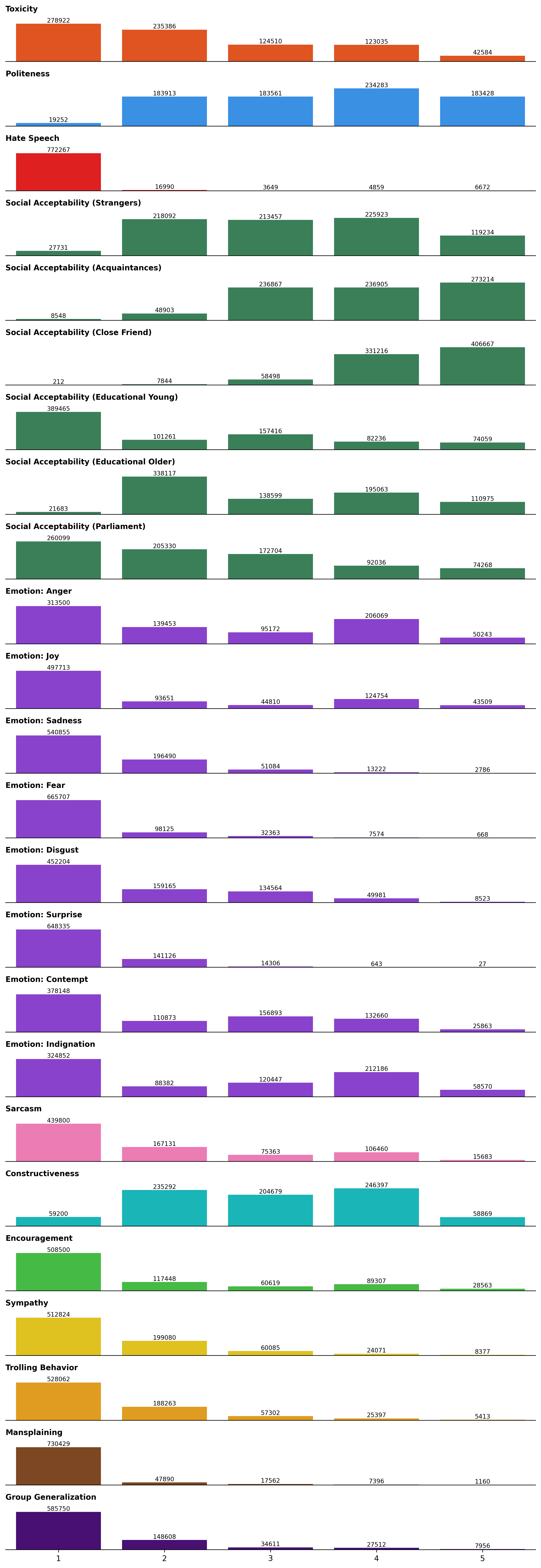}
    \caption{Distribution of AI labels on tasks that were rated from 1 to 5 on a Likert scale.}
    \label{fig:comment-distribution}
\end{figure}

\begin{figure}
    \centering
    \includegraphics[width=1.0\linewidth]{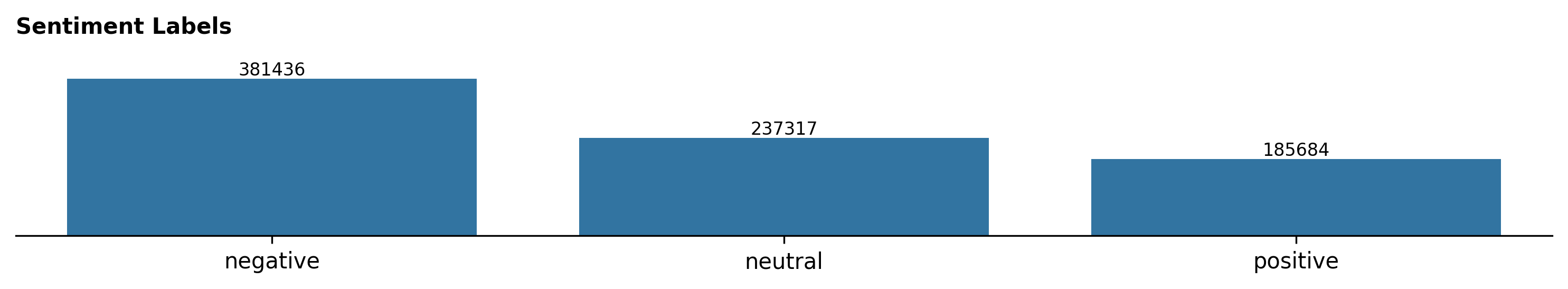}
    \caption{Distribution of AI labels for the sentiment analysis task.}
    \label{fig:sentiment-distribution}
\end{figure}

Before selecting comments for human annotations, we labeled all comments in the 25 different tasks using the GPT-4o mini model. The distribution of labels for each task that was labeled according to a Likert scale is shown in Figure~\ref{fig:comment-distribution} and the distribution of labels in the sentiment task is shown in Figure~\ref{fig:sentiment-distribution}. For sentiment analysis, we observe a somewhat balanced distribution of labels with over 180,000 labels in each sentiment category. For tasks that were rated on a Likert scale, we see great variability in the label distributions. Some tasks, such as toxicity, social acceptability (teacher to young children in an educational environment, parliament speeches), emotion (anger, contempt, indignation), and constructiveness have a somewhat balanced distribution with a significant number of comments in each label category. Tasks such as politeness and social acceptability (strangers, acquaintances, close friends, teacher to teenagers in an educational environment) are skewed to the right and have few comments rated as not having the property of the task. Other tasks are skewed to the left with few comments having the property. For example, 6,672 comments were labeled as having hate speech with strong agreement. The most problematic tasks were ``surprise'' and ``fear'' with only 27 and 668 comments respectively labeled as having the properties with strong agreement.


Our sampling strategy balanced the need for cross-task analysis with the goal of maximizing dataset diversity. We began by creating a shared evaluation set of 100 comments selected uniformly at random from the full corpus. These comments were set as annotation candidates for all 25 tasks, providing a consistent benchmark for analyzing relationships between different aspects of online discourse, such as how toxicity relates to emotion or constructiveness.

For each task, we then selected an additional 1,100 comments that showed strong signals for that specific behavior based on the LLM's ratings (600 comments rated "5" and 500 rated "1"). To maximize dataset diversity and reduce annotator fatigue, we excluded these task-specific comments from the selection in other tasks. This decision reflects the distinct nature of our annotation tasks – a comment exhibiting strong hate speech, for instance, might be uninformative for tasks like encouragement or constructiveness. By presenting annotators with fresh content for each task, we aimed to maintain their engagement and avoid potential biases from repeated exposure to the same comments. Additionally, since we focus on extreme cases, reusing comments across tasks could lead to redundancy, as comments rated extreme in one dimension often represent neutral or irrelevant cases for other dimensions.

The resulting dataset of comment candidates\footnote{Note that not all comments were fully annotated in all task categories.} for human evaluation contains 1,200 comments per task (100 shared + 1,100 task-specific). While this design limits comprehensive cross-task analysis to the shared set of 100 comments, it provides rich, focused data for developing robust classifiers for each individual task. Future work could explore the possibility of annotating a larger shared set of comments across all tasks, which would enable more comprehensive analysis of task relationships while potentially sacrificing some task-specific coverage.

\subsection{Annotator Statistics}
The dataset comprises annotations from 170 unique annotators with an average age of 37.61 years. The educational background of the annotators is diverse, with the majority holding advanced degrees: 36.5\% have a master's degree, 22.9\% have a bachelor's degree, and 5.9\% have a PhD. The gender distribution is nearly balanced, with 47.6\% male and 49.4\% female annotators, while a small percentage identify as other (2.4\%) or prefer not to say (0.6\%). In terms of participation, there is a notable disparity between the average and median number of annotations per user (113.7 and 27.5 respectively), suggesting that while some annotators contributed extensively, the typical annotator provided a more modest number of annotations.

The recruitment and motivation of crowdworkers for annotation tasks can be a challenge. Most of our participants were recruited through targeted Facebook groups, with advertisements highlighting the potential societal benefits of training models to detect hate speech and toxic online behavior. This framing likely contributed to the relatively high number of annotations in these categories. However, task participation decreased for tasks presented later in the annotation sequence, leading to an uneven number of annotations across tasks and a potential annotator bias in those that had a lower number of total annotations. This suggests that fatigue or prioritization may have influenced the workers’ engagement with certain tasks, particularly those positioned further down the task list. In future work, this issue could be mitigated by randomizing the order in which tasks are presented to each crowd worker, thereby ensuring a more balanced distribution of participation across tasks.

\subsection{Agreement}

Table~\ref{tab:aihumanagreement} presents an overview of the annotation statistics and agreement measures for each task in our study. We report several metrics to provide a comprehensive view of the annotation quality and the performance of our AI model compared to human annotators.

To assess the reliability of the annotations, we calculated Krippendorff's alpha \cite[K's $\alpha$]{krippendorff2018content} for inter-annotator agreement. The results varied considerably across tasks, with some showing strong agreement (e.g., disgust: 0.92, sympathy: 0.83) and others showing weaker agreement (e.g., mansplaining: 0.07, fear: 0.24). This variability suggests that some concepts were more challenging to annotate consistently than others. It may be noted that the instructions for mansplaining were more specific for the human annotators than for GPT-4o mini as they explicitly mentioned that the comment should be from a man to a woman. However, that is often an implicit understanding of the word.

To evaluate the performance of our AI model against human consensus, we computed Cohen's kappa \cite[C's $\kappa$]{cohen1960coefficient} between the AI predictions and the aggregated human labels. The AI model showed moderate to substantial agreement with human annotators on several tasks, including politeness (0.82), social acceptability in educational settings (0.74), and emotion detection for anger and joy (both 0.68). However, the model struggled with more nuanced tasks such as mansplaining (0.17) and sarcasm detection (0.23).

Interestingly, some tasks exhibited a discrepancy between human inter-annotator agreement and AI-human agreement. For instance, the sympathy task had high human agreement (K's $\alpha$ = 0.83) but low AI-human agreement (C's $\kappa$ = 0.24), suggesting that while humans consistently identified sympathy, the AI model had difficulty capturing this concept accurately. However, it should be noted that while certainly a valid translation for ``sympathy'', the Icelandic term ``samúð'' has a tendency to be linked exclusively to condolences made on the occasion of the death of a person's relative or friend. It is therefore conceivable that our human annotators have a more narrow understanding of the word than that used by the AI model.

The sentiment analysis task, which involved a three-way classification, showed moderate agreement both among human annotators (K's $\alpha$ = 0.64) and between the AI and human consensus (C's $\kappa$ = 0.59).

The results highlight the varying degrees of difficulty in annotating different aspects of online discourse. While some tasks, particularly those related to basic emotions and clearly defined concepts, showed high agreement, others involving more nuanced or context-dependent judgments proved more challenging for both human annotators and our AI model. Most of the time, if a task has low inter-annotator agreement, the human-AI agreement will also be low, indicating that concepts like sarcasm and trolling are simply difficult to detect in text. It is, however, interesting to note the cases where inter-annotator agreement is high but human-AI agreement is low. For instance, GPT-4o mini does not seem to have a good grasp of the emotions disgust and surprise.

\begin{table*}[htpb]
\small
\begin{center}
\scalebox{0.85}{ 
\begin{tabular}{lrrrrrr}
\toprule
Task & Count & $A\geq 2$ & AAPC & K's $\alpha$ & C's $\kappa$ \\
\midrule
Emotion disgust & 355 & 32 & 1.09 & 0.86 & 0.53 \\
Social acceptability acquaintances & 395 & 44 & 1.11 & 0.77 & 0.71 \\
Emotion contempt & 342 & 37 & 1.11 & 0.77 & 0.63 \\
Emotion surprise & 359 & 23 & 1.07 & 0.75 & 0.35 \\
Encouragement presence & 448 & 69 & 1.17 & 0.74 & 0.66 \\
Emotion joy & 525 & 127 & 1.28 & 0.69 & 0.69 \\
Emotion sadness & 381 & 49 & 1.13 & 0.68 & 0.50 \\
Emotion anger & 547 & 106 & 1.22 & 0.67 & 0.72 \\
Politeness & 749 & 286 & 1.49 & 0.67 & 0.80 \\
Social acceptability educational young & 448 & 68 & 1.16 & 0.65 & 0.73 \\
Group generalization presence & 585 & 156 & 1.32 & 0.64 & 0.62 \\
Social acceptability strangers & 526 & 129 & 1.29 & 0.62 & 0.76 \\
Hate speech presence & 877 & 429 & 1.70 & 0.61 & 0.60 \\
Sentiment & 1099 & 837 & 2.64 & 0.61 & 0.61 \\
Social acceptability educational older & 390 & 51 & 1.14 & 0.58 & 0.75 \\
Constructiveness & 464 & 71 & 1.16 & 0.53 & 0.53 \\
Sympathy & 460 & 63 & 1.15 & 0.53 & 0.25 \\
Toxicity & 981 & 585 & 2.01 & 0.52 & 0.65 \\
Social acceptability close friend & 381 & 33 & 1.09 & 0.44 & 0.36 \\
Emotion fear & 384 & 48 & 1.13 & 0.43 & 0.60 \\
Social acceptability parliament & 404 & 58 & 1.15 & 0.39 & 0.51 \\
Trolling behavior & 511 & 111 & 1.25 & 0.38 & 0.47 \\
Emotion indignation & 354 & 26 & 1.08 & 0.33 & 0.56 \\
Sarcasm & 507 & 89 & 1.19 & 0.29 & 0.26 \\
Mansplaining & 572 & 136 & 1.29 & 0.28 & 0.21 \\
\midrule
Average & 521.76 & 146.52 & 1.30 & 0.58 & 0.56 \\
\bottomrule
\end{tabular}
}
\end{center}
\caption{Overview of the annotations by task. The count column represents the number of comments annotated for each task. The $A\geq 2$ represents the number of comments with two or more annotations. AAPC represents the average number of non-skipped annotations per comment. K's $\alpha$ corresponds to Krippendorff's $\alpha$ amongst the human annotators in the task. Finally, C's $\kappa$ refers to Cohen's $\kappa$ between the AI model and a human consensus label. The last row shows the total for the first two numerical columns and a macro average for the other columns.}\label{tab:aihumanagreement}
\end{table*}

\section{Discussion}

The gold standard, human annotated Hotter and Colder dataset is relatively small. While its main purpose is to serve as validation for the AI-labeled silver dataset, it can also be used as training data for few-shot learning models. The silver dataset offers considerable flexibility, supporting the training of models for individual tasks, such as the automated detection of hate speech. However, the utility of both datasets extends beyond single-task applications. Multi-Task Learning (MTL) allows a model to tackle multiple tasks simultaneously, drawing on shared representations and insights across tasks to improve overall performance. In sentiment analysis, for example, an MTL framework enables a more nuanced understanding of human communication. \citet{tan2023sentiment} demonstrate how sarcasm detection can significantly enhance the performance of sentiment analysis models, particularly in identifying negative sentiment in sarcastic contexts. Our results indicate that sarcasm detection remains a challenge, likely contributing to the suboptimal performance of the model in the sentiment analysis task. Given that Icelandic humor often relies on sarcasm, this cultural factor may explain some of the difficulties the model encounters in this task. Consequently, it is plausible that an Icelandic sentiment analysis model would benefit from an MTL approach, particularly one that integrates sarcasm detection as a complementary task.


When working with multilingual LLMs, cultural norms exhibited by the model might not always match those of the country in question~\cite{meadows2024localvaluebench}. Rather, these models reflect the cultural, legal, and ideological values of their creators. \citet{tao2024cultural} showcased that GPT-4o mini generally mirrors values that are commonly found in English-speaking and Protestant European countries. 
While this cultural bias may not be inherently problematic for our purposes, it could lead to reduced agreement between human annotators and AI models in culture-specific annotations. For instance, ethical alignment performed during model training may influence the model's ability to judge appropriateness in social contexts. A model might consistently classify toxic or hateful comments as unacceptable, even when human annotators might tolerate such comments in specific contexts, such as in private conversations among friends or informal parliamentary discourse. These nuances in cultural and ethical standards may hinder the model’s performance in tasks requiring a deep understanding of social norms and context.

On the flip side of the coin, Hotter and Colder additionally offers invaluable insight into the sociolinguistic patterns of a small online community. Future research will i.a. include an analysis of how discourse changes in liaison with current events, which communities are most affected by toxic behaviors and hate speech, and the characteristics of toxic users.


\section{Conclusion}
This study presents Hotter and Colder, a dataset annotated for 25 tasks that examine various types of online behaviors. By leveraging both AI-based silver labeling and human-in-the-loop gold labeling, we ensure a comprehensive approach to annotating toxic behaviors, emotions, sentiments, and more in Icelandic blog comments. This dual-phase annotation methodology enabled the identification of rare but critical instances of harmful speech while maintaining high annotator agreement across a variety of tasks.

The introduction of a Multi-Task Learning framework as a future direction holds promise for improving the detection of complex phenomena, such as sarcasm, which remains a challenge for both AI models and human annotators, particularly in culturally specific contexts. By integrating tasks such as sarcasm detection with sentiment analysis, future models may achieve greater accuracy and nuanced understanding in detecting various forms of harmful and toxic speech.


Hotter and Colder lays the foundation for future work on mitigating bias and improving ethical alignment in AI models for Icelandic, hopefully fostering safer and more inclusive online environments.

\section{Ethical Considerations}
In our efforts to recruit crowd workers, we appealed mostly to their desire to fight against toxic online behavior and to help aid in the eventual creation of an automatic content moderation tool. Recruiting crowd workers without offering compensation for their work can be considered problematic. We acknowledge that this fact is the likely cause for the relatively unbalanced annotations across tasks. In our case, participants were informed during the recruitment process that a random participant would receive a prize. However, with sufficient financing, it would be more sustainable and fair towards the participants to pay each annotator based on their contributions.

Furthermore, the content in question is inherently problematic in nature. We instructed users to only participate in tasks they were comfortable with and warned them about potential triggers in the content. One user pointed out to us that only being able to label one task at a time for each comment can be unpleasant. For instance, a comment can both have a positive sentiment and exhibit hate speech at the same time. Furthermore, several of the comments will likely be of mixed valence but the annotators were only able to label the comments on either a binary or a 3-class labeling scheme. We acknowledge this limitation. 

We also acknowledge that we studied gender from a binary perspective. We decided to go for that approach since non-binary gender identities can be significantly harder to infer based on usernames. We encourage future researchers to be more inclusive in their research.

We acknowledge the significant computational resources and associated carbon footprint involved in using GPT-4o mini to analyze 800,000 comments, especially given the final dataset size of approximately 12,000 annotated comments. While this approach may appear computationally inefficient at first glance, it served a crucial methodological purpose: identifying rare but important cases of problematic content that would have been extremely resource-intensive to locate through random sampling alone. Traditional approaches requiring human annotators to sift through hundreds of thousands of comments to find relatively rare instances of hate speech or other harmful content would have been prohibitively expensive and potentially more damaging to annotator wellbeing through extended exposure to toxic content. Future work should explore more environmentally sustainable approaches, such as using smaller, task-specific models for initial filtering or developing more efficient sampling strategies that could achieve similar results with less computational overhead.

\section{Ethics Approval}
Running this study as a crowdsourcing project was approved by the ethics board of the University of Iceland (SHV2024-080).

\section{Acknowledgements}
This work was supported by The Ludvig Storr Trust
no. LSTORR2023-93030 and The Icelandic Language Technology Programme.

\bibliographystyle{acl_natbib}
\bibliography{references}

\end{document}